\documentclass{article}

\usepackage{microtype}
\usepackage{graphicx}
\usepackage{subfigure}
\usepackage{booktabs} 
\usepackage[acronym]{glossaries}

\makeglossaries
\newacronym{CNN}{CNN}{Convolutional Neural Network}
\newacronym{ANN}{ANN}{Artificial Neural Network}
\newacronym{EC}{EC}{Evolutionary Computation}
\newacronym{DENSER}{DENSER}{Deep Evolutionary Network Structured Representation}
\newacronym{NE}{NE}{NeuroEvolution}
\newacronym{EA}{EA}{Evolutionary Algorithm}
\newacronym{GA}{GA}{Genetic Algorithm}
\newacronym{GE}{GE}{Grammatical Evolution}
\newacronym{CFG}{CFG}{Context-Free Grammar}
\newacronym{ML}{ML}{Machine Learning}
\newacronym{CoDeepNEAT}{CoDeepNEAT}{Coevolution DeepNEAT}
\newacronym{SANE}{SANE}{Symbiotic, Adaptive Neuro-Evolution}
\newacronym{NEAT}{NEAT}{NeuroEvolution of Augmenting Topologies}
\newacronym{CGP}{CGP}{Cartesian Genetic Programming}
\newacronym{SGE}{SGE}{Structured Grammatical Evolution}
\newacronym{NN-SGE}{NN-SGE}{Neural Networks Structured Grammatical Evolution}
\newacronym{FFNN}{FFNN}{Feed-Forward Neural Network}
\newacronym{BNF}{BNF}{Backus-Naur Form}
\newacronym{GSNN}{GSNN}{Grammatical Swarm Neural Networks}
\newacronym{EANN}{EANN}{Evolutionary Artificial Neural Network}
\newacronym{GP}{GP}{Genetic Programming}
\newacronym{GGP}{GGP}{Grammar-based Genetic Programming}
\newacronym{DSGE}{DSGE}{Dynamic Structured Grammatical Evolution}
\newacronym{DL}{DL}{Deep Learning}
\newacronym{TWEANN}{TWEANN}{Topology and Weight Evolving Artificial Neural Network}
\newacronym{BP}{BP}{BackPropagation}
\newacronym{CoSyNE}{CoSyNE}{Cooperative Synapse NeuroEvolution}
\newacronym{MLP}{MLP}{Multilayer Perceptron}

\usepackage{hyperref}
\usepackage{multirow}

\usepackage[accepted]{icml2018}

\icmltitlerunning{DENSER: Deep Evolutionary Network Structured Representation}

\begin{document}

\twocolumn[
\icmltitle{DENSER: Deep Evolutionary Network Structured Representation}



\icmlsetsymbol{equal}{*}

\begin{icmlauthorlist}
\icmlauthor{Filipe Assun\c{c}\~{a}o}{uc}
\icmlauthor{Nuno Louren\c{c}o}{uc}
\icmlauthor{Penousal Machado}{uc}
\icmlauthor{Bernardete Ribeiro}{uc}
\end{icmlauthorlist}

\icmlaffiliation{uc}{CISUC, Department of Informatics Engineering, University of Coimbra, Coimbra, Portugal}

\icmlcorrespondingauthor{Filipe Assun\c{c}\~{a}o}{fga@dei.uc.pt}

\icmlkeywords{Machine Learning, ICML}

\vskip 0.3in
]



\printAffiliationsAndNotice{\icmlEqualContribution} 

\begin{abstract}
Deep Evolutionary Network Structured Representation (DENSER) is a novel approach to automatically design \glspl{ANN} using Evolutionary Computation. The algorithm not only searches for the best network topology (e.g., number of layers, type of layers), but also tunes hyper-parameters, such as, learning parameters or data augmentation parameters. The automatic design is achieved using a representation with two distinct levels, where the outer level encodes the general structure of the network, i.e., the sequence of layers, and the inner level encodes the parameters associated with each layer. The allowed layers and range of the hyper-parameters values are defined by means of a human-readable Context-Free Grammar. DENSER was used to evolve \glspl{ANN} for CIFAR-10, obtaining an average test accuracy of 94.13\%. The networks evolved for the CIFAR-10 are tested on the MNIST, Fashion-MNIST, and CIFAR-100; the results are highly competitive, and on the CIFAR-100 we report a test accuracy of 78.75\%. To the best of our knowledge, our CIFAR-100 results are the highest performing models generated by methods that aim at the automatic design of \glspl{CNN}, and are amongst the best for manually designed and fine-tuned \glspl{CNN}. 
\end{abstract}

\glsresetall

\section{Introduction}
\label{sec:introduction}

The design of \glspl{ANN} usually requires an arduous and iterative trial-and-error process, where various aspects of the networks have to be considered. Practitioners have to decide on the network topology, the specific parameters of each layer, which learning algorithm should be used and its parameters, and the parameterisation of other criteria like the data pre-processing and/or augmentation methods. Such decisions require a high level of expertise, and if not performed with care, we might design models that have a poor performance . This task becomes even more difficult considering that the different decisions that have to be made are not independent from one another. One way to avoid this laborious process is to resort to networks that have already been constructed for a specific task, and have shown a good performance. Nevertheless, these networks tend to be problem specific, and thus, for each dataset and/or task they might not give the best results. Another approach to overcome this challenge is to rely on automatic methods that seek for the design of \glspl{ANN}.

There are several iterative approaches that seek the structured optimisation of \glspl{ANN}. Constructive~\cite{frean1990upstart,parekh2000constructive} methods start from an elementary structure and add nodes or connections until a network structure that is capable of solving the problem emerges. In contrast, Pruning~\cite{reed1993pruning,molchanov2016pruning} methods start from a complex network structure, and at each iteration remove nodes or connections. These methods are based on the theory that small networks generalise more easily; however that is not necessarily true~\cite{sietsma1991creating}. Another limitation of these methods is that often only a single network is being optimised, and consequently, there is a high chance of the search becoming stagnated in a local optima. 

To address the problem of automatic design of \glspl{ANN} we propose \gls{DENSER}, a novel approach for the automatic generation of the topology and hyper-parameters needed to build effective \glspl{ANN}. \gls{DENSER} is a layer-based method. The evolution of each neuron directly allows for a higher degree of freedom which results in greater control over the generated structures. However, when dealing with big datasets where large scale networks are needed, the involved number of neurons and connections make evolution at such a low level unfeasible. This work builds upon our earlier work~\cite{assuncao2017evolving}, with the demonstration that deep \glspl{ANN} generated for specific tasks can generalise to tasks where they have not been trained on during evolution.

The main contributions of this work are:
\begin{itemize}
    \item \gls{DENSER}, a general framework based on evolutionary principles that automatically searches for the adequate structure and parameterisation of large scale deep networks that can have different layer types (e.g., convolutional, pooling, fully-connected), and goals (e.g., classification, regression);
    \item An automatically generated \gls{CNN} that without any prior knowledge is effective on the classification of the CIFAR-10 dataset, with an average test accuracy of $94.13\%$;
    \item The demonstration that \glspl{ANN} evolved with DENSER generalise well. In concrete, the best network whose topology was evolved for the CIFAR-10 dataset reports average accuracy values of 99.70\%, 95.26\%, and 78.75\% on the MNIST, Fashio-MNIST, and CIFAR-100, respectively. To the best of our knowledge, these are the best results reported by methods that automatically design \glspl{CNN}.
\end{itemize}

The best trained models have been released at \url{http://github.com/fillassuncao/denser-models}.

The remainder of the paper is organised as follows. In Section~\ref{sec:related_work} we introduce the concepts for understanding \gls{DENSER}. Next, in Section~\ref{sec:denser}, we describe our novel approach, \gls{DENSER}, which is followed by the conducted set of experiments (reported in Section~\ref{sec:experiments}). To end, in Section~\ref{sec:conclusions}, conclusions are drawn and future work is addressed. 

\section{Related Work}
\label{sec:related_work}

\gls{DENSER} is a \gls{NE} approach, and thus is based on the use of \glspl{EA} to automatically optimise \glspl{ANN}. Next, we detail the principles behind \glspl{EA} and review works that are closely related to ours.

 \begin{figure}[t!]
     \centering
     \includegraphics[width=0.48\textwidth]{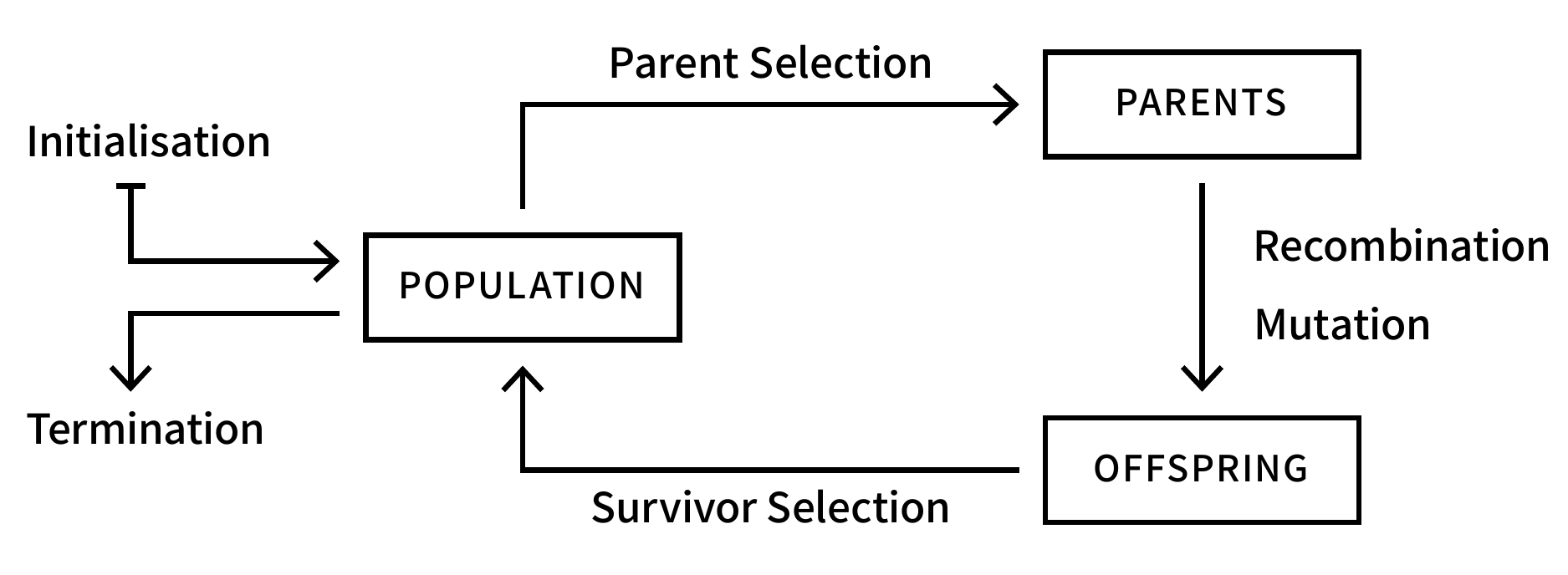}
     \caption{Evolutionary algorithms flow-chart.}
     \label{fig:ea_flowchart}
 \end{figure}

 \subsection{Evolutionary Algorithms}
 \label{sec:ec}

 \glspl{EA} are stochastic search procedures inspired by the principles of natural selection and genetics, that have been successfully applied in optimisation, design and learning problems~\cite{eiben2015}. Historically, there are several variants of \glspl{EA}, but they share the same common underlying idea: the simulation of evolution of a population of artificial individuals by natural selection (proposed by Darwin) via the application of selection, variation operators (typically crossover and mutation), and reproduction. These components are guided by a fitness function that evaluates each individual, measuring the quality of the solution it represents. The application of these components is repeated for several iterations, and over time it is expected that the overall quality of the individuals in the population improves. The process stops when a predetermined termination criterion is met (e.g., when a maximum number of iterations is achieved). Each artificial individual in \glspl{EA} encodes a single candidate solution to the problem being considered. 

 The general flow-chart of a simple EA is shown in Figure~\ref{fig:ea_flowchart}. The main components are: 
 \begin{description}
     \item[Representation] -- defines how the solutions to the considered problem should be encoded. The genetic material used to represent the solutions is known as genotype; the expression of the genotype is the phenotype.
     \item[Evaluation] -- estimates how good a solution is in solving the problem under consideration. Usually it is a mathematical expression and enables the comparison between problem solutions.
     \item[Parent Selection] -- selects, probabilistically, the population individuals (called parents) to participate in the breeding of a new population. 
     \item[Variation Operators] -- create new individuals (offspring) using the parents. These operators are used in a stochastic manner, and are usually divided in two types: crossover, and mutation. Crossover creates variation in the population by taking two, or more individuals, as input, and rearranges their information to create new solutions. Mutation takes one individual as input and slightly modifies it.
     \item[Survivor Selection] -- determines the solutions that proceed to the next iteration of the \gls{EA}. The number of individuals in an EA is typically kept fixed.
 \end{description}

\subsection{NeuroEvolution}
\label{sec:neuroevolution}


\glsreset{NE}\gls{NE}~\cite{floreano2008neuroevolution} is a sub-field of \gls{ML} and \gls{EC} that applies evolutionary methods to the optimisation of \glspl{ANN}. \gls{NE} approaches are commonly grouped according to the aspects of the \glspl{ANN} that they optimise: (i) learning~\cite{radi2003discovering,gomez2008accelerated,morse2016simple}; (ii) topology~\cite{harp1990designing,soltanian2013artificial,rocha2007evolution}; or (iii) both topology and learning~\cite{whitley1990genetic,stanley2002evolving,turner2013cartesian}. 

The vast majority of \gls{NE} works target the evolution of small networks for very specific tasks. With \gls{DENSER} our goal is to evolve large scale networks that can deal with vast amounts of data and challenging tasks. As an example, consider the VGG network~\cite{simonyan2014very}: a 16 to 19 deep \gls{CNN} that is often used for image recognition tasks. The number of neurons and connections involved in a deep architectures usually turns connection~\cite{kitano1990designing,leung2003tuning,fernando2017pathnet} or node-based~\cite{moriarty1997forming,stanley2002evolving,assuncao2017towards} evolutionary methods impractical for discovering high performing networks, due to the large search space that needs to be scanned. Therefore, for evolving deep networks practitioners often resort to layer-based encodings~\cite{jung2006evolutionary,suganuma2017cnns,miikkulainen2017evolving}. For similar reasons, it is unfeasible to directly evolve the weights of the networks, which easily reach the range of thousands, or even millions of parameters; when the training of the networks is optimised using \gls{EC} usually only the hyper-parameters are tuned and the networks trained using gradient-descent algorithms~\cite{miikkulainen2017evolving,suganuma2017cnns}.

Loshchilov and Hutter~\cite{loshchilov2016cma} developed an approach based on Evolutionary Strategies~\cite{hansen2001completely} to optimise the hyper-parameters of deep networks; the tuned parameters are concerned with the topology (e.g., number of filters in the convolution layers, and number of neurons in fully-connected layers) and learning (e.g., batch size, and learning rates). This approach requires the \textit{a-priori} definition of the network structure that may be suitable for solving the problem, and consequently there is no optimisation of the sequence of layers and connections between them. 

The idea of optimising hyper-parameters for deep networks is further extended in \gls{CoDeepNEAT}~\cite{miikkulainen2017evolving}, where the structure of the network is searched combining the ideas behind \gls{SANE}~\cite{moriarty1997forming} and \gls{NEAT}~\cite{stanley2002evolving}. Two populations are evolved in simultaneous: one of modules and another one of blueprints, which specify the modules that should be used. Learning and data augmentation parameters are also optimised. 

A similar approach is proposed in CGP-NN~\cite{suganuma2017cnns}, where \gls{CGP}~\cite{miller2000cartesian} is used in the evolution of the architecture of \glspl{CNN}. However, instead of automatically searching the modules, they are specified by the user, and just their combination and parameterisation is evolved. 

In this work we want to make the automatic generation of deep networks as easy and transparent as possible. That is the reason why we adopt a grammar-based approach. Grammars allow us to specify different network types, such as AutoEncoders or \glspl{CNN}, without the need to change any implementation details. Further, grammar-based methods make it easy to incorporate knowledge, allowing the specification of modules and/or parameters that we may know or suspect that work well on certain problem domains. 

There are several approaches concerned with the evolution of \glspl{ANN} using \gls{GE}~\cite{o2003grammatical}. The vast majority of them focus on the tuning of a single hidden-layer, i.e., on the number of neurons and their connections from the input to the hidden-nodes and from the hidden-nodes to the outputs~\cite{soltanian2013artificial,ahmadizar2015artificial,assuncao2017evolving}. In~\cite{assuncao2017towards} a grammar-based method that is able to evolve networks with more than one hidden-layer is described. Despite theoretically suit to generate deep networks, the involved amount of neurons and connections make the domain space too large to be searched within an acceptable time.
\begin{table}[]
\centering
\caption{Classification accuracy of various \glspl{CNN} on MNIST, Fashion-MNIST, CIFAR-10, and CIFAR-100. Automatic approaches are marked with an *.}
\label{tab:cifar_sota_results}
\resizebox{\columnwidth}{!}{
\begin{tabular}{c|c|c}
\textbf{Approach}     & \textbf{Dataset}                                & \textbf{Accuracy} \\ \hline
\multicolumn{1}{c|}{\cite{simard2003best}} &  \multicolumn{1}{c|}{\multirow{2}{*}{MNIST}}                     &   99.60\%                \\
\multicolumn{1}{c|}{\cite{graham2014fractional}} & \multicolumn{1}{c|}{}                           &  99.68\%                 \\ \hline
\multicolumn{1}{c|}{\cite{AlexNet}} &  \multicolumn{1}{c|}{\multirow{3}{*}{Fashion-MNIST}}                     &   89.90\%                \\
\multicolumn{1}{c|}{\cite{simonyan2014very}} & \multicolumn{1}{c|}{}                           &  93.50\%                 \\ 
\multicolumn{1}{c|}{\cite{ResNet18}} & \multicolumn{1}{c|}{}                           &  94.90\%                 \\ \hline
\multicolumn{1}{c|}{\cite{loshchilov2016cma}*} & \multicolumn{1}{c|}{\multirow{6}{*}{CIFAR-10}}  &  90.70\%                 \\
\multicolumn{1}{c|}{\cite{miikkulainen2017evolving}*} & \multicolumn{1}{c|}{}                           &     92.70\%              \\
\multicolumn{1}{c|}{\cite{snoek2015scalable}*} & \multicolumn{1}{c|}{}                           &   93.63\%                \\
\multicolumn{1}{c|}{\cite{suganuma2017cnns}*} & \multicolumn{1}{c|}{}                           &    94.02\%               \\
\multicolumn{1}{c|}{\cite{real2017large}*} & \multicolumn{1}{c|}{}                           &    95.60\%               \\
\multicolumn{1}{c|}{\cite{graham2014fractional}} & \multicolumn{1}{c|}{}                           &    96.53\%               \\ \hline
\multicolumn{1}{c|}{\cite{snoek2015scalable}*} & \multicolumn{1}{c|}{\multirow{3}{*}{CIFAR-100}} &                   72.60\%\\
\multicolumn{1}{c|}{\cite{graham2014fractional}} & \multicolumn{1}{c|}{}                           &      73.61\%             \\
\multicolumn{1}{c|}{\cite{real2017large}*} & \multicolumn{1}{c|}{}                           &      77.00\%            
\end{tabular}
}
\end{table}

Table~\ref{tab:cifar_sota_results} reports on the performance of different models for the classification of the MNIST, Fashion-MNIST, CIFAR-10 and CIFAR-100. Those models discovered by automatic approaches are marked with an *. We only report results for the datasets that are used in the conducted experiments.

\section{Deep Evolutionary Network Structured Representation}
\label{sec:denser}

\glsreset{DENSER} \gls{DENSER} is a novel representation that combines the basic principles of two \glspl{EA}: \glspl{GA}~\cite{mitchell1998introduction} and \gls{DSGE}~\cite{assuncao2017towards}. \gls{DSGE} is a variant of \gls{GE}~\cite{o2003grammatical,lourencco2016unveiling}, and thus a form of Genetic Programming (GP)~\cite{koza1992genetic}. Whilst in the vanilla version of GP the solutions are encoded via a syntax-tree, in \gls{GE} approaches the candidate solutions are encoded using variable length integer arrays, which represent derivations of a user-defined \gls{CFG}. Formally, a \gls{CFG} is a tuple $G=(N,T,S,P)$, where $N$ is a non-empty set of non-terminal symbols, $T$ is a non-empty set of terminal symbols, $S \in N$ is the starting symbol, and $P$ is the set of production rules of the form $A ::= \alpha$, with $A \in N$ and $\alpha \in (N \cup T)^*$. $N$ and $T$ are disjoint. Each grammar $G$ defines a language $L(G)$ composed by all sequences of terminal symbols that can be derived from the starting symbol: $L(G)= \{ w:\, S\overset{*} {\Rightarrow} w,\, w \in T^*\}$. Next, we further detail each component of \gls{DENSER}.

\begin{figure}[t!]
    {\scriptsize
    \begin{align*}
        {<}\text{features}{>} ::= & \, {<}\text{convolution}{>} \\
                   & \, | \, {<}\text{pooling}{>}\\
        {<}\text{convolution}{>} ::= & \, \text{layer:conv} \, \text{[num-filters,int,1,32,256]} \\
                   & \, \text{[filter-shape,int,1,1,5]} \, \text{[stride,int,1,1,3]} \\
                   & \, {<}\text{padding}{>} \, {<}\text{activation}{>} \, {<}\text{bias}{>}\\
                   & \, {<}\text{batch-normalisation}{>} \, {<}\text{merge-input}{>}  \\
        {<}\text{batch-normalisation}{>} ::= & \, \text{batch-normalisation:True} \\
                   & \, | \, \text{batch-normalisation:False}\\
        {<}\text{merge-input}{>} ::= & \, \text{merge-input:True} \\
                   & \, | \, \text{merge-input:False}\\
        {<}\text{pooling}{>} ::= & \, {<}\text{pool-type}{>} \, \text{[kernel-size,int,1,1,5]} \\
                                 & \, \text{[stride,int,1,1,3]} \, {<}\text{padding}{>} \\
        {<}\text{pool-type}{>} ::= & \, \text{layer:pool-avg} \\
                   & \, | \, \text{layer:pool-max}\\
        {<}\text{padding}{>} ::= & \, \text{padding:same} \\
                   & \, | \, \text{padding:valid}\\
        {<}\text{classification}{>} ::= & \, {<}\text{fully-connected}{>} \\
        {<}\text{fully-connected}{>} ::= & \, \text{layer:fc} \, {<}\text{activation}{>} \, \\
                                         & \text{[num-units,int,1,128,2048]} \, {<}\text{bias}{>} \\
        {<}\text{activation}{>} ::= & \, \text{act:linear} \\
                   & \, | \, \text{act:relu}\\
                   & \, | \, \text{act:sigmoid}\\
        {<}\text{bias}{>} ::= & \, \text{bias:True} \\
                   & \, | \, \text{bias:False}\\
        {<}\text{softmax}{>} ::= & \, \text{layer:fc} \, \text{act:softmax} \, \text{num-units:10} \, \text{bias:True} \\
        {<}\text{learning}{>} ::= & \, \text{learning:gradient-descent} \, \text{[lr,float,1,0.0001,0.1]}
    \end{align*}}
    \vspace{-20pt}\caption{Example grammar for the encoding of \glspl{CNN}.}
    \label{fig:grammar}
\end{figure}

\subsection{Representation}
\label{sec:representation}

Each solution encodes an~\gls{ANN} by means of an ordered sequence of feedforward layers and their respective parameters; the learning, data augmentation, and any other hyper-parameters can be encoded with each individual too. The representation of the candidate solution is made at two different levels: 
\begin{description}
    \item[\gls{GA} Level] -- encodes the macro structure of the networks and is responsible for representing the sequence of layers that later serves as an indicator to the grammatical starting symbol. It requires the definition of the allowed structure of the networks, i.e., the valid sequence of layers. For example, for evolving \glspl{CNN} the following \gls{GA} structure may be specified: $[$(features, 1, 10), (classification, 1, 2), (softmax, 1, 1), (learning, 1, 1)$]$, where each tuple indicates the valid starting symbols, and the minimum and maximum number of times they can be used. Using the grammar of Figure~\ref{fig:grammar}, this \gls{GA} structure can evolve networks with up to 10 convolution or pooling layers, followed by up to 2 fully-connected layers, and the classification layer softmax, that usually has a number of outputs that matches the number of problem classes; the learning tuple is responsible for codifying the parameters that should be used to train the network.
    \item[\gls{DSGE} Level] -- encodes the parameters associated to a layer. The parameters and their allowed values or ranges are codified in the grammar that must be defined by the user. Looking at the grammar of Figure~\ref{fig:grammar}, for the pooling layers we tune the kernel size, the stride, and the type of padding. The same exercise can be made to the remaining layers defined in the grammar. The parameters can have closed values (e.g., the padding that can be only valid or same), or can assume a value in an integer or real interval.
\end{description}

\begin{figure}[t!]
    \centering
    \includegraphics[width=\linewidth]{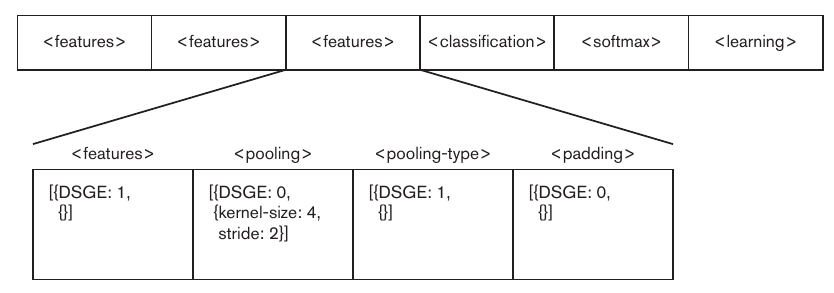}
    \caption{Example of the genotype of a candidate solution that encodes a \gls{CNN}.}
    \label{fig:genotype}
    \vspace{10pt}
    \includegraphics[width=0.6\linewidth]{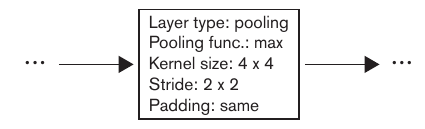}
    \caption{Phenotype corresponding to the only layer specified in Figure~\ref{fig:genotype}.}
    \label{fig:phenotype}
\end{figure}

The novel combination of a \gls{GA} with \gls{DSGE} enables a \mbox{two-fold} gain: (i) the \gls{GA} level encapsulates the genetic material of each layer, making it easier to apply the variation operators (described next); and (ii) the \gls{DSGE} makes the approach easily generalisable, as it is only needed to change the grammar to enable the evolution of different types of networks, or of networks to solve different tasks; it also facilitates the incorporation of domain specific knowledge. 

An example of the genotype is shown in Figure~\ref{fig:genotype}. This example is based on the grammar of Figure~\ref{fig:grammar} and on the above detailed \gls{GA} structure. Figure~\ref{fig:phenotype} depicts the phenotype corresponding to the layer which has the \gls{DSGE} genotype detailed in Figure~\ref{fig:genotype}.

\subsection{Variation Operators}
\label{sec:genetic_operators}

To promote the evolution of the solutions we design variation operators that act on the two levels of the genotype.

\begin{figure}[t!]
    \centering
    \includegraphics[width=\linewidth]{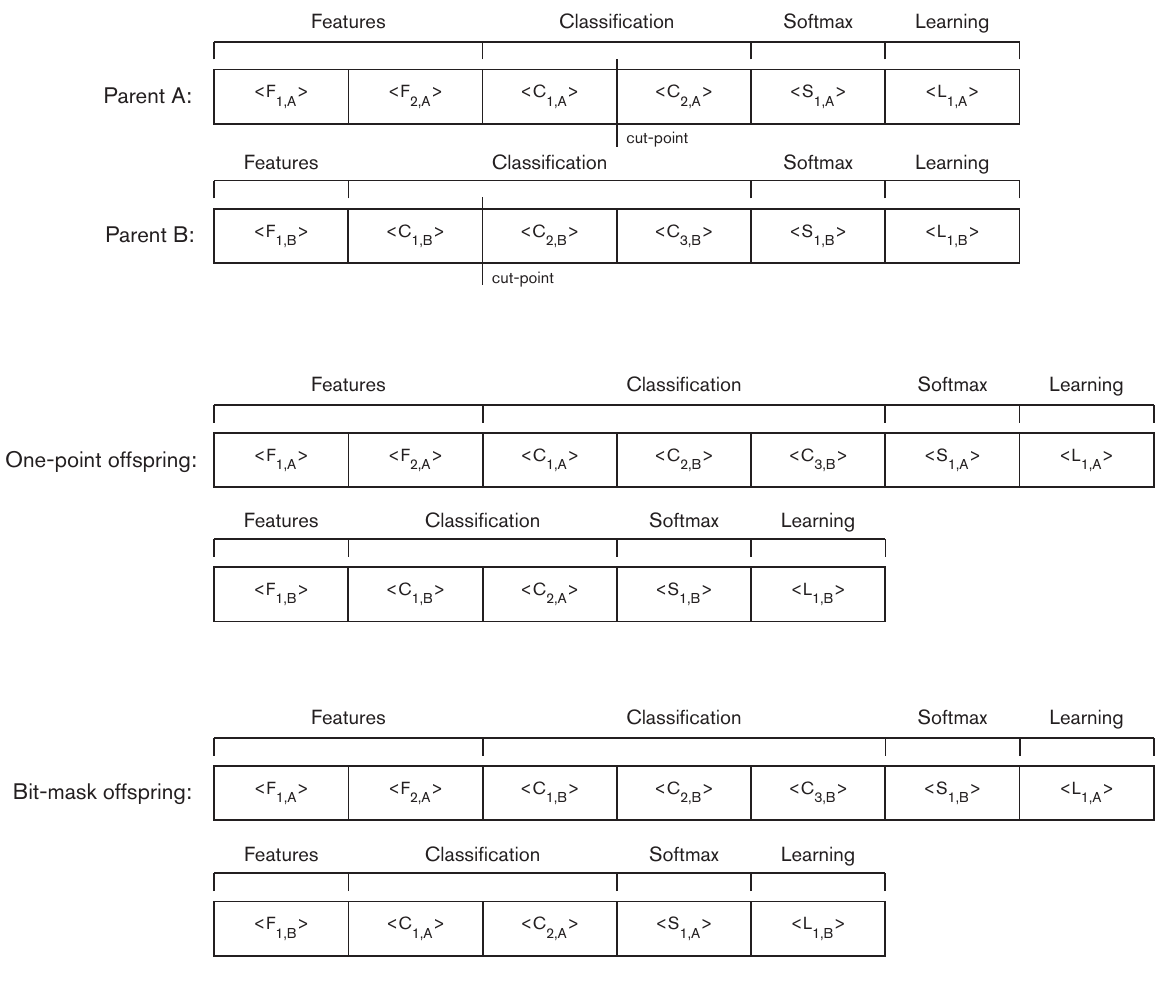}
    \vspace{-7pt}\caption{Example of the introduced crossover operators. The example focuses on the \gls{GA} level of the genotype. For the bit-mask crossover the mask is 1001, which is associated to the features, classification, softmax and learning modules, respectively.}
    \label{fig:crossover}
\end{figure}

\subsection*{Crossover}

We designed two crossover operators, both of them based on the premise that the genetic material is encapsulated, which facilitates the exchange between individuals. In the context of this work a module does not refer to a set of layers that can be replicated multiple times, but is rather the set of layers that belongs to the same \gls{GA} structure index.


To exchange layers within the same module we use a one-point crossover. Imagining that we are evolving bitstrings, if the parents are $111|000$ and $101|010$, and $|$ represents the cut-point, the offspring results from changing the genetic material delimited by the cut-point, i.e., $111010$ and $101000$. The same module in different individuals may vary in size; the cut-point is generated at random taking into account the smallest module. 

To exchange modules between two parents we use a \mbox{bit-mask} crossover. In the bit-mask crossover, as the name suggests first we need to create a mask of bits of the size of the number of number of codons (i.e., modules) that are to be exchanged (in the case of the above \gls{GA} structure, 4). Then the first offspring is created by copying a codon from the first parent if the bit is $1$, and if the bit is $0$ the codon is copied from the second parent. The opposite is done for the second offspring: we copy from the first parent if the bit is $0$, and from the second parent if the bit is $1$.

An example of the application of the crossover operators is depicted in Figure~\ref{fig:crossover}.

\subsection*{Mutation}

We develop two sets of mutation operators that act upon the \gls{GA} and \gls{DSGE} levels, respectively. The mutations on the \gls{GA} level are three:
\begin{description}
    \item[Add layer] -- generates a new layer randomly, subjected to the constraints of the module where it will be placed.
    \item[Replicate layer] -- selects a module and copies an existing layer to another position of the module; the copy is made by reference, meaning that any change in the parameters of the layer is propagated to the copies.
    \item[Remove layer] -- deletes a random layer from a module, without violating the minimum number of layers. 
\end{description}

The mutations on the \gls{DSGE} level are two: 
\begin{description}
    \item[Grammatical mutation] -- as in standard \gls{DSGE}, an expansion possibility is replaced by another valid one;
    \item[Integer/float mutation] -- an integer/float block is replaced by a new one. For integers we generate new integers at random; for floats a Gaussian perturbation is used.
\end{description}

\subsection{Evaluation}
\label{sec:evaluation}

In \gls{DENSER}, as in the majority of layer-based \gls{NE} approaches, we only allow the evolution of the learning \mbox{hyper-parameters}. Therefore, to evaluate the candidate solutions we must train them on the task that we are trying to solve. In this work we are going to perform object recognition tasks, using the CIFAR-10~\cite{krizhevsky2009learning}. The quality of the solutions is measured based on the accuracy metric. The training and evaluation are performed by mapping the solutions generated by \gls{DENSER}, i.e., strings that result from the grammatical derivation of the defined grammar, into a  Keras~\cite{chollet2015keras} model running on top of TensorFlow~\cite{abadi2016tensorflow}, and then executed on a GPU. Each network is trained during $10$ epochs; in the end we return the best accuracy on the validation set.

While in traditional \gls{ML} approaches we often only need two dataset partitions (train and test), in \gls{NE} we need three:
\begin{description}
\item[Train --] used to train the network using the defined or evolved learning parameters;
\item[Validation --] used to evaluate the performance of the network during evolution, i.e., the accuracy on the validation set is used as the fitness metric;
\item[Test --] kept aside from the evolutionary process, and used to evaluate the performance of the best models on unseen data, so we can understand the generalisation ability.
\end{description}
If we define no test set and only use two partitions it is impossible to test the evolved models on data that is not presented to the model during evolution. Thus the reported results would be biased. Cross-validation is not applied due to the associated computational cost. Moreover, we followed the same data augmentation method reported by \cite{suganuma2017cnns}, which includes: padding, horizontal flips, and random crops.

\section{Experiments}
\label{sec:experiments}

To analyse the performance of \gls{DENSER} we perform experiments on the generation of \glspl{CNN} for the classification of the CIFAR-10 dataset. Then, to assess the generalisation and scalability ability of the networks that are evolved using \gls{DENSER} we take the best model found for the classification of the CIFAR-10 dataset and re-train it on the MNIST, Fashion-MNIST, and  CIFAR-100.

\subsection{Datasets}
\label{sec:dataset}

The CIFAR-10~\cite{krizhevsky2009learning} is composed by 32 $\times$ 32 RGB colour images, in a total of 60000 instances; each real-world image contains one of the following objects: airplane, automobile, bird, cat, deer, dog, frog, horse, ship, or truck. MNIST~\cite{lecun1998gradient} is composed by 60000 grayscale images, each with a handwritten digit (between 0 and 9). MNIST is known for being an easy to solve problem; therefore we also investigate the performance of the \glspl{CNN} on Fashion-MNIST~\cite{xiao2017/online}: similar to MNIST (i.e., 28 $\times$ 28 grayscale images), but where the digits digits are replace by fashion clothing items from Zalando's: t-shirt/top, trouser, pullover, dress, coat, sandal, shirt, sneaker, bag, or ankle boot.

The previous datasets are all composed by instances of 10 independent classes; to test the ability of DENSER to solve problems with more classes we use CIFAR-100~\cite{krizhevsky2009learning}, that has the same 60000 instances as CIFAR-10, but where the instance labels have a higher resolution. The goal in all tasks is to maximise the accuracy of the recognition.

\begin{table}[t!]
    \centering
    \small
    \caption{Experimental parameters.}\vspace{5pt}
    \label{tab:exp_parameters}
    \begin{tabular}{c|c}
        \textbf{Evolutionary Engine Parameter} & \textbf{Value} \\ \hline
        Number of runs & 10 \\ 
        Number of generations & 100 \\ 
        Population size & 100 \\
        Crossover rate & 70\% \\
        Mutation rate & 30\% \\
        Tournament size & 3 \\
        Elite size & 1\% \\ 
        \textbf{Dataset Parameter} & \textbf{Value} \\ \hline
        Train set & 42500 instances \\
        Validation set & 7500 instances \\
        Test set & 10000 instances \\
        \textbf{Training Parameter} & \textbf{Value} \\ \hline
        Number of epochs & 10 \\
        Loss & Categorical Cross-entropy \\
        Batch size & 125\\
        Learning rate & 0.01\\
        Momentum & 0.9\\
        \textbf{Data Augmentation Parameter} & \textbf{Value} \\ \hline
        Padding & 4 \\
        Random crop & 4 \\
        Horizontal flipping & 50\% \\
    \end{tabular}
\end{table}

\subsection{Experimental Setup}
\label{sec:experimental_setup}

The experimental parameters used are detailed in Table~\ref{tab:exp_parameters}. We use the grammar of Figure~\ref{fig:grammar} and the following \gls{GA} structure: $[$(features, 1, 30), (classification, 1, 10), (softmax, 1, 1)$]$. This way our search space encompasses \glspl{CNN} that have up to $40$ hidden-layers: at most $30$ convolution or pooling layers followed by up to $10$ fully-connected layers. The minimum number of layers of the networks is 3.

The train of a \gls{CNN} is a computationally expensive task. For that reason, during the evolutionary process we only perform $10$ epochs for each network. In the end of each evolutionary run, we take the best network and perform longer trains, with $400$ epochs and the same learning rate policy. For these extended trains, the train and validation sets used in the evolutionary runs are merged, so that more data is available for learning.

\begin{figure*}[t!]
    \centering
    \includegraphics[width=0.7\textwidth]{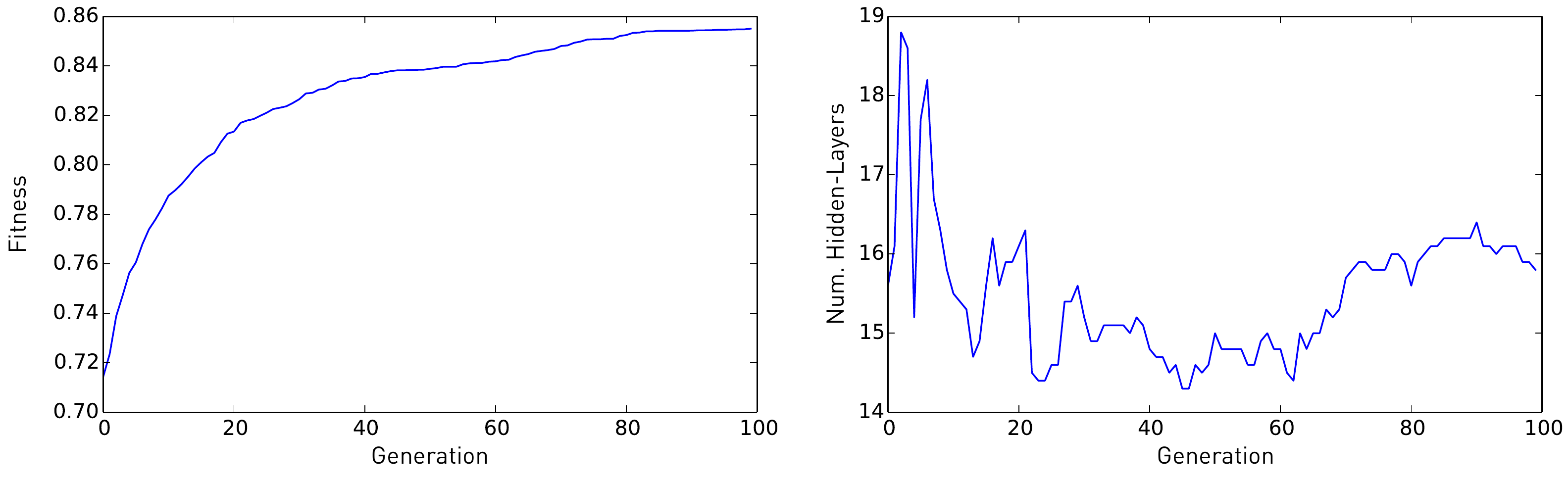}
    \vspace{-7pt}\caption{Evolution of the fitness (left) and number of hidden-layers (right) of the best individuals across generations. Results are averages of $10$ independent runs.}
    \label{fig:fit_evolution}
\end{figure*}

\subsection{Evolution of CNNs for the CIFAR-10 Dataset}
\label{sec:cifar10_results}

The fitness evolution of the best networks across the generations is depicted in Figure~\ref{fig:fit_evolution}. An inspection of the results shows that the performance of the networks is steadily increasing, and evolution converges around the 80th generation. It is possible to see that a change in behaviour occurs around the 60th generation: before, the increase in fitness is followed by a decrease in the number of hidden-layers; after, the increase in fitness is accompanied by an increase in the number of hidden-layers. To support this analysis we compute the Pearson correlation between the fitness values of the best individuals, and the average number of layers, per generation. Until the 60th generation the Pearson correlation reports a coefficient of $-0.7166$ (moderate negative correlation); after the 60th generation the correlation coefficient is $0.9204$ (strong positive correlation).

The change in behaviour around the 60th generation is easily explainable. At first, the candidate solutions are randomly generated and have approximately 22.44 hidden-layers (population average). These initial solutions have their parameters generated at random, and thus, it is highly unlikely that a stochastic parameterisation of 22 layers will have any meaningful results. As time passes, the networks decrease in complexity and their parameters are tuned, until a point where to increase the performance of the networks more layers are necessary (60th generation).  

Once the evolutionary process is over, the best network of each run is re-trained $5$ times during $400$ epochs. The best network of each run is selected according to the accuracy values on the validation set. All the accuracy values reported from this point onwards are averages of $5$ independent trains; this is done because the initial weights are different. This training conditions lead to an average classification accuracy on the test set of $88.41\%$.

The comparison of this result with the ones reported by other approaches (check Table~\ref{tab:cifar_sota_results}) seems to indicate that \gls{DENSER} performs worse. However, different approaches use different training policies, data augmentation, or prediction techniques. For example, \cite{snoek2015scalable} for each instance of the test set generate $100$ augmented images, and the prediction of the image class is based on the class that has the maximum average confidence on the $100$ augmented images. By doing this the average accuracy on the test set increases to $89.93\%$.

To check if it is still possible to enhance the performance of the best networks we test a different learning rate policy. In \cite{suganuma2017cnns} the authors use a varying learning rate policy: it starts at $0.01$; on the 5th epoch it is increased to $0.1$; by the 250th epoch it is decreased to $0.01$; and finally at the 375th it is reduced to $0.001$; Nesterov momentum is used. The previous changes lead to an average test accuracy of $92.51\%$, not using data augmentation on the test set. Applying data augmentation to the test set, further increases the average accuracy to $93.29\%$.

Recall that all the previous results are averages of the $10$ best networks, one from every of the performed evolutionary runs. Each network is trained 5 times. The average accuracy on the test set of the network that has the highest accuracy on 5 trains is $93.37\%$ (not using data augmentation on the test set), or $94.13\%$ (using data augmentation of the test set). These results are with the learning rate policy of \cite{suganuma2017cnns}. 

\begin{figure}[t!]
    \centering
    \includegraphics[width=0.5\textwidth]{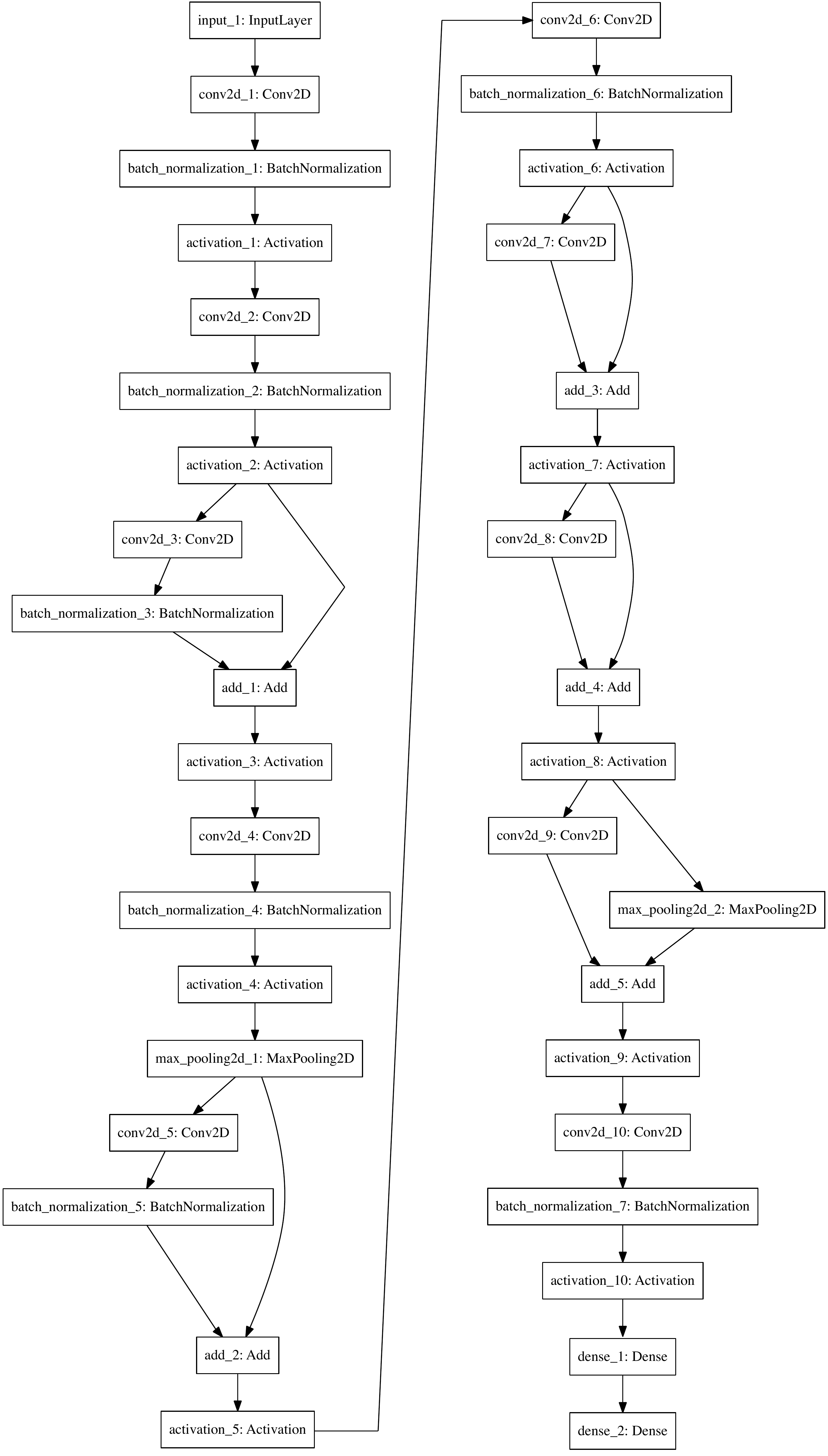}
    \vspace{-7pt}\caption{Topology of the \gls{CNN} that reports the best results.}
    \label{fig:best_network}
\end{figure}

Focusing on the structure of the best evolved networks, the most puzzling characteristic is the importance and number of the fully-connected layers. On average, the best networks have $2.2$ fully-connected layers that are placed before the softmax layer. We conducted preliminary experiments where some of the fully-connected layers were removed; results show that the performance of the networks generated. This outcome of the evolution is remarkable, since the majority of hand-designed networks only have one dense layer, i.e., the classification layer. Figure~\ref{fig:best_network} depicts the network that reports the best accuracy on the test set over all the conducted experiments.

\subsection{MNIST, Fashion-MNIST, and CIFAR-100}
\label{sec:cifar100_results}

To test the generalisation and scalability of the evolved networks we take the best network generated by \gls{DENSER} on the CIFAR-10 and apply it to the classification of the MNIST, Fashion-MNIST, and CIFAR-100. Except for the CIFAR-100, all the others have the same number of classes as CIFAR-10; in order for the best \gls{CNN} to work on the CIFAR-100 we just need to adapt the softmax layer to have 100 units instead of 10. 

To re-train the best networks we use the training policy that in the previous experiments reported the best results, i.e., we train during 400 epochs, with the same learning rate of CGP-CNN~\cite{suganuma2017cnns}: a varying learning rate that starts at $0.01$, and that on the 5th, 250th and 375th epochs changes to $0.1$, $0.01$, and $0.001$, respectively. We do 5 independent trains of the network, and as above we report average results. 

On the MNIST the \gls{CNN} attains an average classification accuracy of 99.65\% without data augmentation over the test set, and 99.70\% when data augmentation is used. Nowadays MNIST is known to be an easy to solve problem, where simple \glspl{MLP} attain very high accuracies, close to 99\%~\cite{simard2003best}. That is the reason why we have also opted for testing the networks found with DENSER on the Fashion-MNIST~\cite{xiao2017/online}, where we obtain average accuracies of 94.23\% (and 94.70\%), not using (and using) data augmentation on the test set.

The previous experiments prove that the \glspl{CNN} evolved by \gls{DENSER} generalise: the networks are evolved for one dataset and are robust enough to correctly classify others. Further, they do so without the need to change any hyper-parameter (nor from the structure, nor learning). Nonetheless, we still do not know if the networks scale, i.e., if despite evolved for distinguishing between 10 independent classes, they are able to separate a larger number of classes. To investigate this we use the CIFAR-100.

On the CIFAR-100, the average accuracy is of 73.32\% when not augmenting the test data; the average accuracy increases to 74.94\% when using data augmentation. The results on CIFAR-100 demonstrate that the \gls{CNN} that are not evolved for solving problems with a higher dimensionality are able to do so, with state-of-the-art performances.

The train of each network is stochastic; the initial conditions (i.e., initial weights) are different and they are trained using different instances of the dataset, because of the data augmentation methodology. Thus, and in order to further improve the results, we investigate if the approach proposed in \cite{graham2014fractional} to test the performance of the fractional max-pooling increases the performance reported by our network. In brief words, instead of a single network, an ensemble is used, where each network that is part of the ensemble is the result of an independent train of the network. 

We apply the previous methodology to all datasets; the accuracies on the MNIST, Fashion-MNIST, and CIFAR-100 are 99.70\%, 95.11\%, and 77.51\%, respectively. That is, the results of ensembling the 5 trains are at least equal, and often superior to the reported by just a single network. Moreover, and instead of forming ensembles with the same network structure we also test the performance of building an ensemble formed by the two best network topologies found by \gls{DENSER}, similarly to what is done in~\cite{real2017large}. This way, we obtain accuracy values of 99.70\%, 95.26\%, and 78.75\%, respectively.

\subsection{Discussion}
\label{sec:discussion}

In the above sections we have demonstrated that DENSER is able to successfully evolve \glspl{CNN} for the CIFAR-10, and that the networks that are found by evolution generalise to the MNIST, and Fashion-MNIST, and scale to the CIFAR-100. 

The test classification accuracy of the fittest \gls{CNN} on the CIFAR-10 is 93.29\%, and on MNIST, Fashion-MNIST, and CIFAR-100 are of 99.70\%, 94.70\%, and 74.94\%, respectively. The results of the ensemble formed by the two fittest networks are of 94.13\%, 99.70\%, 95.26\%, 78.75\%, respectively for the CIFAR-10, MNIST, Fashion-MNIST, and CIFAR-100. Therefore, the ensemble provides results that are consistently superior (or at least equal) to the single highest performing \gls{CNN}.

The comparison with the state-of-the-art is carried out based on Table~\ref{tab:cifar_sota_results}. The best results for CIFAR-10, MNIST, Fashion-MNIST, and CIFAR-100 are 96.53\%, 99.68\%, 94.90, and 77\%, respectively. Based on all the previous results it is possible to conclude that only considering the single fittest \gls{CNN}, DENSER is highly competitive with the state-of-the-art results on MNIST and Fashion-MNIST. Notwithstanding, not all results reported on Table~\ref{tab:cifar_sota_results} are generated by automatic search methods. On the CIFAR-10, and CIFAR-100 if constraining to automatic methods, the best results are 95.60\%, and 77\%, respectively, and are both obtained by \cite{real2017large}, which use ensembles. If we compare their results with our ensembling ones we note that on CIFAR-10 they perform slightly better (difference of 1.47\%), but on CIFAR-100 \gls{DENSER} is the method that stands out (difference of 1.75\%). The reason for that is related with the a-priori knowledge provided to the systems, that enables \cite{real2017large} to find superior results on CIFAR-10; however, we note that the results that DENSER reports on the CIFAR-100 are not the result of evolution, and are an outcome of generalisation, robustness, and scalability.

\section{Conclusions and Future Work}
\label{sec:conclusions}

The current paper describes DENSER: a general-purpose framework for automatically searching for deep networks. The representation of the candidate solutions is made at two different levels: (i) the GA-level that encapsulates the genetic material, thus facilitating the application of the genetic operators; and (ii) the DSGE-level that makes the approach easily generalisable to deal with different problems, and network types or layers. 

We test DENSER on the evolution of \glspl{CNN} for the CIFAR-10 dataset. The results show the effectiveness of the method, which is capable to generate \glspl{CNN} with a classification accuracy of up to 94.13\%. To analyse the generalisation and scalability of the \glspl{CNN} found by DENSER we take the best networks found during evolution and apply them to the classification of the MNIST, Fashion-MNIST, and CIFAR-100. Without further evolution we are able to classify the datasets with high performances, competitive with the state-of-the-art. The most striking result is the performance of 78.75\% on the CIFAR-100, which to the best of our knowledge sets a new state-of-the-art on automatically discovered \glspl{CNN}. 

Notwithstanding the obtained results, there is still room for improvement. We need to come up with better ways for assessing the performance of the networks. The networks are only being trained for 10 epochs, which may bias search towards networks that train fast; however, longer trains mean longer evaluation cycles. The resolution of this limitation makes it possible to optimise the learning parameters more effectively, and also introduce other types of layers (e.g., dropout), that have no effect on such short trains.

\section*{Acknowledgements}
\noindent This work is partially funded by: Funda\c{c}\~ao para a Ci\^encia e Tecnologia (FCT), Portugal, under the grant SFRH/BD/114865/2016. We would also like to thank NVIDIA for providing us Titan X GPUs.


\begin{thebibliography}{45}
\providecommand{\natexlab}[1]{#1}
\providecommand{\url}[1]{\texttt{#1}}
\expandafter\ifx\csname urlstyle\endcsname\relax
  \providecommand{\doi}[1]{doi: #1}\else
  \providecommand{\doi}{doi: \begingroup \urlstyle{rm}\Url}\fi

\bibitem[Abadi et~al.(2016)Abadi, Barham, Chen, Chen, Davis, Dean, Devin,
  Ghemawat, Irving, Isard, et~al.]{abadi2016tensorflow}
Abadi, Mart{\'\i}n, Barham, Paul, Chen, Jianmin, Chen, Zhifeng, Davis, Andy,
  Dean, Jeffrey, Devin, Matthieu, Ghemawat, Sanjay, Irving, Geoffrey, Isard,
  Michael, et~al.
\newblock Tensorflow: A system for large-scale machine learning.
\newblock In \emph{OSDI}, volume~16, pp.\  265--283, 2016.

\bibitem[Ahmadizar et~al.(2015)Ahmadizar, Soltanian, AkhlaghianTab, and
  Tsoulos]{ahmadizar2015artificial}
Ahmadizar, Fardin, Soltanian, Khabat, AkhlaghianTab, Fardin, and Tsoulos,
  Ioannis.
\newblock Artificial neural network development by means of a novel combination
  of grammatical evolution and genetic algorithm.
\newblock \emph{Engineering Applications of Artificial Intelligence},
  39:\penalty0 1--13, 2015.

\bibitem[Assun{\c{c}}\~ao et~al.(2017{\natexlab{a}})Assun{\c{c}}\~ao,
  Louren{\c{c}}o, Machado, and Ribeiro]{assuncao2017evolving}
Assun{\c{c}}\~ao, Filipe, Louren{\c{c}}o, Nuno, Machado, Penousal, and Ribeiro,
  Bernardete.
\newblock Evolving the topology of large scale deep neural networks.
\newblock In \emph{EuroGP}, volume~1. Springer, 2017{\natexlab{a}}.

\bibitem[Assun{\c{c}}\~ao et~al.(2017{\natexlab{b}})Assun{\c{c}}\~ao,
  Louren{\c{c}}o, Machado, and Ribeiro]{assuncao2017towards}
Assun{\c{c}}\~ao, Filipe, Louren{\c{c}}o, Nuno, Machado, Penousal, and Ribeiro,
  Bernardete.
\newblock Towards the evolution of multi-layered neural networks: A dynamic
  structured grammatical evolution approach.
\newblock In \emph{Proceedings of the Genetic and Evolutionary Computation
  Conference}, GECCO '17, pp.\  393--400, New York, NY, USA,
  2017{\natexlab{b}}. ACM.
\newblock ISBN 978-1-4503-4920-8.
\newblock \doi{10.1145/3071178.3071286}.
\newblock URL \url{http://doi.acm.org/10.1145/3071178.3071286}.

\bibitem[Chollet et~al.(2015)]{chollet2015keras}
Chollet, Fran\c{c}ois et~al.
\newblock Keras.
\newblock \url{https://github.com/keras-team/keras}, 2015.

\bibitem[Eiben \& Smith(2015)Eiben and Smith]{eiben2015}
Eiben, Agoston and Smith, James.
\newblock \emph{Introduction to evolutionary computing}.
\newblock Natural computing series. Springer, Berlin, Heidelberg, Paris, 2015.
\newblock ISBN 3-540-40184-9.

\bibitem[Fernando et~al.(2017)Fernando, Banarse, Blundell, Zwols, Ha, Rusu,
  Pritzel, and Wierstra]{fernando2017pathnet}
Fernando, Chrisantha, Banarse, Dylan, Blundell, Charles, Zwols, Yori, Ha,
  David, Rusu, Andrei~A., Pritzel, Alexander, and Wierstra, Daan.
\newblock Pathnet: Evolution channels gradient descent in super neural
  networks.
\newblock \emph{arXiv preprint arXiv:1701.08734}, 2017.

\bibitem[Floreano et~al.(2008)Floreano, D{\"u}rr, and
  Mattiussi]{floreano2008neuroevolution}
Floreano, Dario, D{\"u}rr, Peter, and Mattiussi, Claudio.
\newblock Neuroevolution: from architectures to learning.
\newblock \emph{Evolutionary Intelligence}, 1\penalty0 (1):\penalty0 47--62,
  2008.

\bibitem[Frean(1990)]{frean1990upstart}
Frean, Marcus.
\newblock The upstart algorithm: A method for constructing and training
  feedforward neural networks.
\newblock \emph{Neural Computation}, 2\penalty0 (2):\penalty0 198--209, 1990.

\bibitem[Gomez et~al.(2008)Gomez, Schmidhuber, and
  Miikkulainen]{gomez2008accelerated}
Gomez, Faustino, Schmidhuber, J{\"u}rgen, and Miikkulainen, Risto.
\newblock Accelerated neural evolution through cooperatively coevolved
  synapses.
\newblock \emph{Journal of Machine Learning Research}, 9\penalty0
  (May):\penalty0 937--965, 2008.

\bibitem[Graham(2014)]{graham2014fractional}
Graham, Benjamin.
\newblock Fractional max-pooling.
\newblock \emph{arXiv preprint arXiv:1412.6071}, 2014.

\bibitem[Hansen \& Ostermeier(2001)Hansen and Ostermeier]{hansen2001completely}
Hansen, Nikolaus and Ostermeier, Andreas.
\newblock Completely derandomized self-adaptation in evolution strategies.
\newblock \emph{Evolutionary computation}, 9\penalty0 (2):\penalty0 159--195,
  2001.

\bibitem[Harp et~al.(1990)Harp, Samad, and Guha]{harp1990designing}
Harp, Steven~A, Samad, Tariq, and Guha, Aloke.
\newblock Designing application-specific neural networks using the genetic
  algorithm.
\newblock In \emph{Advances in neural information processing systems}, pp.\
  447--454, 1990.

\bibitem[He et~al.(2016)He, Zhang, Ren, and Sun]{ResNet18}
He, Kaiming, Zhang, Xiangyu, Ren, Shaoqing, and Sun, Jian.
\newblock Deep residual learning for image recognition.
\newblock In \emph{Proceedings of the IEEE conference on computer vision and
  pattern recognition}, pp.\  770--778, 2016.

\bibitem[Jung \& Reggia(2006)Jung and Reggia]{jung2006evolutionary}
Jung, Jae-Yoon and Reggia, James~A.
\newblock Evolutionary design of neural network architectures using a
  descriptive encoding language.
\newblock \emph{IEEE transactions on evolutionary computation}, 10\penalty0
  (6):\penalty0 676--688, 2006.

\bibitem[Kitano(1990)]{kitano1990designing}
Kitano, Hiroaki.
\newblock Designing neural networks using genetic algorithms with graph
  generation system.
\newblock \emph{Complex systems}, 4\penalty0 (4):\penalty0 461--476, 1990.

\bibitem[Koza(1992)]{koza1992genetic}
Koza, John~R.
\newblock \emph{Genetic programming: on the programming of computers by means
  of natural selection}, volume~1.
\newblock MIT press, 1992.

\bibitem[Krizhevsky \& Hinton(2009)Krizhevsky and
  Hinton]{krizhevsky2009learning}
Krizhevsky, Alex and Hinton, Geoffrey.
\newblock Learning multiple layers of features from tiny images, 2009.

\bibitem[Krizhevsky et~al.(2012)Krizhevsky, Sutskever, and Hinton]{AlexNet}
Krizhevsky, Alex, Sutskever, Ilya, and Hinton, Geoffrey~E.
\newblock Imagenet classification with deep convolutional neural networks.
\newblock In \emph{Advances in neural information processing systems}, pp.\
  1097--1105, 2012.

\bibitem[LeCun et~al.(1998)LeCun, Bottou, Bengio, and
  Haffner]{lecun1998gradient}
LeCun, Yann, Bottou, L{\'e}on, Bengio, Yoshua, and Haffner, Patrick.
\newblock Gradient-based learning applied to document recognition.
\newblock \emph{Proceedings of the IEEE}, 86\penalty0 (11):\penalty0
  2278--2324, 1998.

\bibitem[Leung et~al.(2003)Leung, Lam, Ling, and Tam]{leung2003tuning}
Leung, Frank Hung-Fat, Lam, Hak-Keung, Ling, Sai-Ho, and Tam, Peter Kwong-Shun.
\newblock Tuning of the structure and parameters of a neural network using an
  improved genetic algorithm.
\newblock \emph{IEEE Transactions on Neural networks}, 14\penalty0
  (1):\penalty0 79--88, 2003.

\bibitem[Loshchilov \& Hutter(2016)Loshchilov and Hutter]{loshchilov2016cma}
Loshchilov, Ilya and Hutter, Frank.
\newblock {CMA-ES} for hyperparameter optimization of deep neural networks.
\newblock \emph{arXiv preprint arXiv:1604.07269}, 2016.

\bibitem[Louren{\c{c}}o et~al.(2016)Louren{\c{c}}o, Pereira, and
  Costa]{lourencco2016unveiling}
Louren{\c{c}}o, Nuno, Pereira, Francisco~B, and Costa, Ernesto.
\newblock Unveiling the properties of structured grammatical evolution.
\newblock \emph{Genetic Programming and Evolvable Machines}, 17\penalty0
  (3):\penalty0 251--289, 2016.

\bibitem[Miikkulainen et~al.(2017)Miikkulainen, Liang, Meyerson, Rawal, Fink,
  Francon, Raju, Navruzyan, Duffy, and Hodjat]{miikkulainen2017evolving}
Miikkulainen, Risto, Liang, Jason, Meyerson, Elliot, Rawal, Aditya, Fink, Dan,
  Francon, Olivier, Raju, Bala, Navruzyan, Arshak, Duffy, Nigel, and Hodjat,
  Babak.
\newblock Evolving deep neural networks.
\newblock \emph{arXiv preprint arXiv:1703.00548}, 2017.

\bibitem[Miller \& Thomson(2000)Miller and Thomson]{miller2000cartesian}
Miller, Julian~F and Thomson, Peter.
\newblock Cartesian genetic programming.
\newblock In \emph{European Conference on Genetic Programming}, pp.\  121--132.
  Springer, 2000.

\bibitem[Mitchell(1998)]{mitchell1998introduction}
Mitchell, Melanie.
\newblock \emph{An introduction to genetic algorithms}.
\newblock MIT press, 1998.

\bibitem[Molchanov et~al.(2016)Molchanov, Tyree, Karras, Aila, and
  Kautz]{molchanov2016pruning}
Molchanov, Pavlo, Tyree, Stephen, Karras, Tero, Aila, Timo, and Kautz, Jan.
\newblock Pruning convolutional neural networks for resource efficient transfer
  learning.
\newblock \emph{arXiv preprint arXiv:1611.06440}, 2016.

\bibitem[Moriarty \& Miikkulainen(1997)Moriarty and
  Miikkulainen]{moriarty1997forming}
Moriarty, David~E. and Miikkulainen, Risto.
\newblock Forming neural networks through efficient and adaptive coevolution.
\newblock \emph{Evolutionary Computation}, 5\penalty0 (4):\penalty0 373--399,
  1997.

\bibitem[Morse \& Stanley(2016)Morse and Stanley]{morse2016simple}
Morse, Gregory and Stanley, Kenneth~O.
\newblock Simple evolutionary optimization can rival stochastic gradient
  descent in neural networks.
\newblock In \emph{Proceedings of the 2016 on Genetic and Evolutionary
  Computation Conference}, pp.\  477--484. ACM, 2016.

\bibitem[O'Neill \& Ryan(2003)O'Neill and Ryan]{o2003grammatical}
O'Neill, Michael and Ryan, Conor.
\newblock Grammatical evolution.
\newblock In \emph{Grammatical Evolution}, pp.\  33--47. Springer, 2003.

\bibitem[Parekh et~al.(2000)Parekh, Yang, and Honavar]{parekh2000constructive}
Parekh, Rajesh, Yang, Jihoon, and Honavar, Vasant.
\newblock Constructive neural-network learning algorithms for pattern
  classification.
\newblock \emph{IEEE Transactions on Neural Networks}, 11\penalty0
  (2):\penalty0 436--451, 2000.

\bibitem[Radi \& Poli(2003)Radi and Poli]{radi2003discovering}
Radi, Amr and Poli, Riccardo.
\newblock Discovering efficient learning rules for feedforward neural networks
  using genetic programming.
\newblock In \emph{Recent advances in intelligent paradigms and applications},
  pp.\  133--159. Springer, 2003.

\bibitem[Real et~al.(2017)Real, Moore, Selle, Saxena, Suematsu, Le, and
  Kurakin]{real2017large}
Real, Esteban, Moore, Sherry, Selle, Andrew, Saxena, Saurabh, Suematsu,
  Yutaka~Leon, Le, Quoc, and Kurakin, Alex.
\newblock Large-scale evolution of image classifiers.
\newblock \emph{arXiv preprint arXiv:1703.01041}, 2017.

\bibitem[Reed(1993)]{reed1993pruning}
Reed, Russell.
\newblock Pruning algorithms-a survey.
\newblock \emph{IEEE Transactions on Neural Networks}, 4\penalty0 (5):\penalty0
  740--747, 1993.

\bibitem[Rocha et~al.(2007)Rocha, Cortez, and Neves]{rocha2007evolution}
Rocha, Miguel, Cortez, Paulo, and Neves, Jos{\'e}.
\newblock Evolution of neural networks for classification and regression.
\newblock \emph{Neurocomputing}, 70\penalty0 (16):\penalty0 2809--2816, 2007.

\bibitem[Sietsma \& Dow(1991)Sietsma and Dow]{sietsma1991creating}
Sietsma, Jocelyn and Dow, Robert~JF.
\newblock Creating artificial neural networks that generalize.
\newblock \emph{Neural networks}, 4\penalty0 (1):\penalty0 67--79, 1991.

\bibitem[Simard et~al.(2003)Simard, Steinkraus, Platt, et~al.]{simard2003best}
Simard, Patrice~Y, Steinkraus, David, Platt, John~C, et~al.
\newblock Best practices for convolutional neural networks applied to visual
  document analysis.
\newblock In \emph{ICDAR}, volume~3, pp.\  958--962, 2003.

\bibitem[Simonyan \& Zisserman(2014)Simonyan and Zisserman]{simonyan2014very}
Simonyan, Karen and Zisserman, Andrew.
\newblock Very deep convolutional networks for large-scale image recognition.
\newblock \emph{arXiv preprint arXiv:1409.1556}, 2014.

\bibitem[Snoek et~al.(2015)Snoek, Rippel, Swersky, Kiros, Satish, Sundaram,
  Patwary, Prabhat, and Adams]{snoek2015scalable}
Snoek, Jasper, Rippel, Oren, Swersky, Kevin, Kiros, Ryan, Satish, Nadathur,
  Sundaram, Narayanan, Patwary, Mostofa, Prabhat, Mr, and Adams, Ryan.
\newblock Scalable bayesian optimization using deep neural networks.
\newblock In \emph{International Conference on Machine Learning}, pp.\
  2171--2180, 2015.

\bibitem[Soltanian et~al.(2013)Soltanian, Tab, Zar, and
  Tsoulos]{soltanian2013artificial}
Soltanian, Khabat, Tab, Fardin~Akhlaghian, Zar, Fardin~Ahmadi, and Tsoulos,
  Ioannis.
\newblock Artificial neural networks generation using grammatical evolution.
\newblock In \emph{Electrical Engineering (ICEE), 2013 21st Iranian Conference
  on}, pp.\  1--5. IEEE, 2013.

\bibitem[Stanley \& Miikkulainen(2002)Stanley and
  Miikkulainen]{stanley2002evolving}
Stanley, Kenneth~O. and Miikkulainen, Risto.
\newblock Evolving neural networks through augmenting topologies.
\newblock \emph{Evolutionary computation}, 10\penalty0 (2):\penalty0 99--127,
  2002.

\bibitem[Suganuma et~al.(2017)Suganuma, Shirakawa, and Nagao]{suganuma2017cnns}
Suganuma, Masanori, Shirakawa, Shinichi, and Nagao, Tomoharu.
\newblock A genetic programming approach to designing convolutional neural
  network architectures.
\newblock In \emph{Proceedings of the Genetic and Evolutionary Computation
  Conference}, GECCO '17, pp.\  497--504, New York, NY, USA, 2017. ACM.
\newblock ISBN 978-1-4503-4920-8.
\newblock \doi{10.1145/3071178.3071229}.
\newblock URL \url{http://doi.acm.org/10.1145/3071178.3071229}.

\bibitem[Turner \& Miller(2013)Turner and Miller]{turner2013cartesian}
Turner, Andrew~James and Miller, Julian~Francis.
\newblock Cartesian genetic programming encoded artificial neural networks: a
  comparison using three benchmarks.
\newblock In \emph{Proceedings of the 15th annual conference on Genetic and
  evolutionary computation}, pp.\  1005--1012. ACM, 2013.

\bibitem[Whitley et~al.(1990)Whitley, Starkweather, and
  Bogart]{whitley1990genetic}
Whitley, Darrell, Starkweather, Timothy, and Bogart, Christopher.
\newblock Genetic algorithms and neural networks: Optimizing connections and
  connectivity.
\newblock \emph{Parallel computing}, 14\penalty0 (3):\penalty0 347--361, 1990.

\bibitem[Xiao et~al.(2017)Xiao, Rasul, and Vollgraf]{xiao2017/online}
Xiao, Han, Rasul, Kashif, and Vollgraf, Roland.
\newblock Fashion-MNIST: a novel image dataset for benchmarking machine
  learning algorithms, 2017.

\end{thebibliography}
\bibliographystyle{icml2018}

\end{document}